\documentclass[10pt,twocolumn]{article}

\usepackage[a4paper,top=1in,bottom=1in,left=0.7in,right=0.7in,columnsep=0.3in]{geometry}
\usepackage{graphicx}
\usepackage{amsmath}
\usepackage{amssymb}
\usepackage{booktabs}
\usepackage{hyperref}
\usepackage{url}
\usepackage{xcolor}
\usepackage{natbib}           
\usepackage{caption}
\usepackage{subcaption}
\usepackage[normalem]{ulem}

\title{Relightable Gaussian Splatting for Virtual Production Using Image-Based Illumination}

\author{
  Adrian Azzarelli\thanks{University of Bristol, UK. \texttt{a.azzarelli@bristol.ac.uk}} \and
  Nantheera Anantrasirichai \and
  James Pollock\thanks{Lux Aeterna, Bristol, UK. \texttt{james@lavfx.com}} \and
  David R. Bull\thanks{University of Bristol, UK}
}
\date{}

\begin{document}

\maketitle

\begin{abstract}
In-camera visual effects in virtual production (VP) use LED walls to provide both background imagery and image-based lighting. While this enables on-set compositing, it tightly couples lighting to background and scene appearance, limiting flexibility for downstream editing. In addition, inverse rendering conventionally relies on a physically-based rendering framework that jointly estimates 3D geometry and lighting, using environment maps. However, these maps are typically low-resolution and assume far-field lighting. In VP, with near-field and high-resolution image-based lighting, this can lead to inaccurate results as well introducing complexities in the editing process.  To address these issues, we propose a VP-specific framework for 3D reconstruction and relighting using Gaussian Splatting. Our framework leverages the known background imagery to condition the relighting process. This avoids relying on environment maps and reduces compositing to a simple background-image editing task. To realize our framework, we introduce a process (and associated dataset) that captures real VP scenes under varying background content and illumination conditions. This data is used to decompose a 3D scene into fixed appearance and variable lighting components. The variable lighting process simulates light transport by parameterizing each primitive with a learnable UV coordinate, intensity value and resolution modifier. Using mipmaps, these parameters directly sample the background texture in image space - implicitly capturing reflections and refractions without reliance on physically-based rendering processes. Combined with the fixed appearance component, this allows us to render relit scenes using a Gaussian Splatting rasterizer. Compared to baselines, our approach achieves higher-quality 3D reconstruction and controllable relighting. The method is efficient (<3 GB RAM, <5 GB VRAM, <2 hours training, ~35 FPS) and supports rendering useful arbitrary output variables including depth, lighting intensity, lighting color, and unlit renders. A subjective study is also presented that clearly demonstrates the practical benefits of our framework for photorealistic relighting in post production.
\end{abstract}
\section{Introduction}\label{sec: introduction}
In recent years, virtual production (VP) has emerged as a widely adopted alternative to conventional matting techniques (e.g., green-background matting) for compositing real foregrounds objects with virtual backgrounds \cite{chen2024impact}. By rendering background content on large LED walls, VP enables in-camera compositing during filming while naturally providing scene illumination. However, this tight coupling between background imagery and lighting at capture time limits post-production flexibility. In particular, deviations from the desired creative outcome typically necessitate costly and logistically complex re-shoots, and/or post production, due to the reliance on specialized stages, equipment, and personnel. This paper addresses this limitation by enabling the post-capture modification of both the LED background imagery and the associated scene lighting.

Research on background matting using LED displays has partially addressed this challenge using 2-D relighting processes \cite{peers2009compressive, wang2009kernel, sengupta2021light, chuang2000environment, matusik2002image}. However, these may require specialized scanning equipment \cite{chuang2000environment, matusik2002image}, and only operate in 2D image space and therefore cannot synthesize novel 3D views or camera trajectories.

In contrast, we approach this problem using 3D Gaussian Splatting (GS), which enables efficient multi-view reconstruction and high-quality novel view synthesis. For film production, this allows camera motion and shot composition to be designed in post-production, thus overcoming the need for physical rigs such as dollies or jibs during capture. More broadly, novel view synthesis enables the rendering of arbitrary output variables (AOVs) including depth maps and unlit renders, which are valuable for downstream visual effects (VFX) workflows. Our goal is therefore to combine the benefits of 3D scene reconstruction with LED-based background matting to enable shot design, background replacement, and lighting modification in post-production for static VP scenes.
\begin{figure*}
    \centering
    \includegraphics[width=\linewidth]{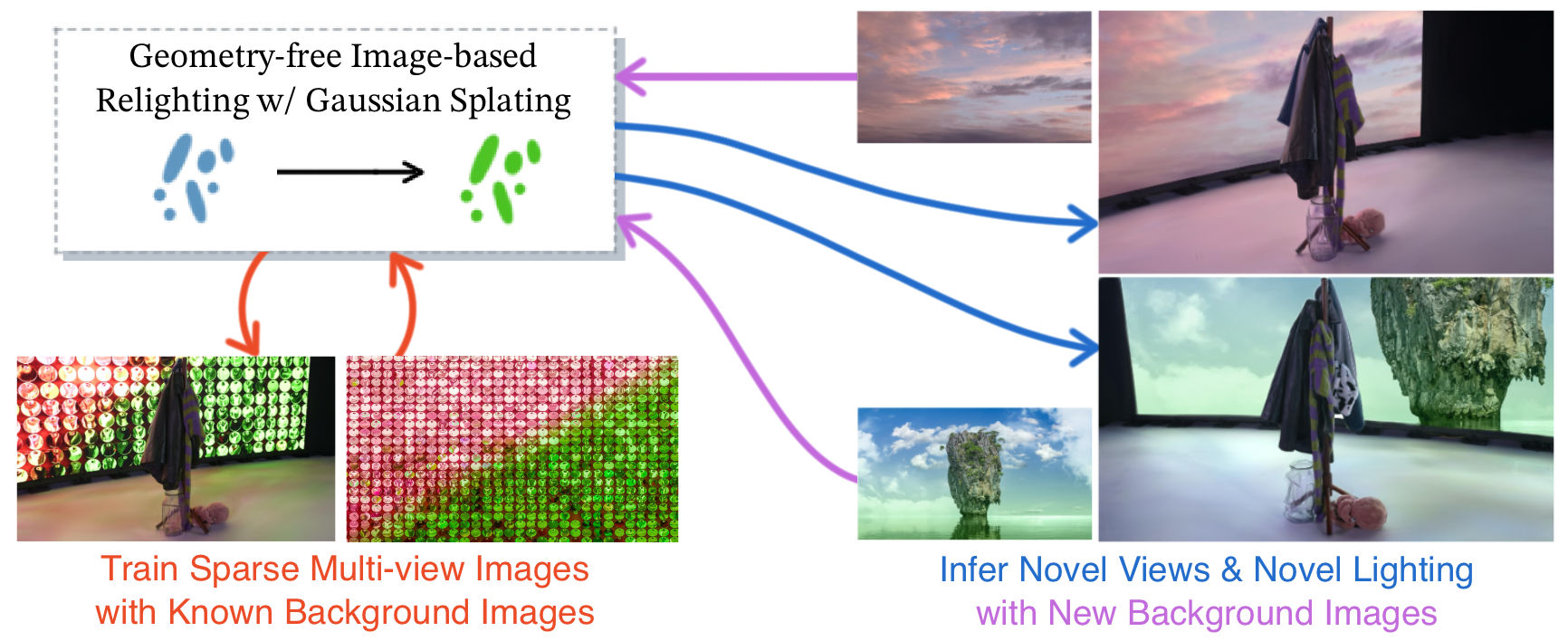}
    \caption{Overview of the 3D reconstruction and image-based relighting problem for virtual production use cases.}

    \label{fig: introduction}
\end{figure*}

Recent GS research has explored relighting via inverse rendering (IR) \cite{bi2024gs3,wu2025deferredgs,liang2024gs,gao2024relightable}. These approaches have attempted to recover geometry and material properties, often parameterized using physically-based shading functions, like the Disney bidirectional reflectance distribution function (BRDF) \cite{gao2024relightable,wu2025deferredgs,guo2024prtgs}. In these approaches, scene illumination is typically represented and learned using environment maps that are captured using spherical or cubic projections and sampled using ray-tracing techniques. During training, these maps can help disentangle lighting from object appearance, and during inference they enable editing scene lighting. However, the standard environment-map formulation assumes distant (far-field) illumination. This assumption is however, violated in VP settings where the LED walls act as near-field light sources, leading to incorrect lighting effects and poor reconstruction results due to mismatches in geometry.

IR approaches also introduce practical limitations. Many rely on GPU-specific ray-tracing frameworks such as NVIDIA OptiX, restricting portability \cite{moenne20243d, wu20253dgut}. Furthermore, robust surface normals are required for ray-tracing, which in turn necessitate ground-truth depth data or auxiliary geometry models \cite{guo2024prtgs, li2025tsgs,zhu2025gaussian}; GS alone can not reliably recover surface geometry, particularly for reflective or transparent materials. Consequently, these methods increase both capture and/or reconstruction complexity, as well as computational cost.

To address these limitations, we propose a geometry-implicit relighting framework for GS scenes capturing VP stages. Rather than relying on environment maps, our method leverages the known background imagery to model and learn the scene illumination. Specifically, we assume that each Gaussian primitive can be associated with a surface-reflected viewport ray that intersects the LED wall's image space. Therefore, each Gaussian is parameterized by a learned a UV coordinate that samples the background image directly in pixel space, enabling efficient approximation of incident illumination without explicit ray-tracing.

To account for occlusions or reflected rays that do not intersect the LED wall, we introduce a per-Gaussian exposure parameter that attenuates the sampled lighting contribution, producing a zero-valued residual when no valid intersection occurs. Adding this residual to each Gaussian's base color decomposes scene lighting into fixed and variable components, enabling relighting through direct manipulation of the background texture — without explicit geometry, light transport simulation, or custom CUDA implementations. This makes our method well suited for VP pipelines where capture simplicity, portability, and computational efficiency are critical.

To enable background matting, we create foreground masks for each camera view. These allow us to remove background pixels during training and prune Gaussians corresponding to the LED wall. After training, the original background imagery is reintroduced by projecting it onto a simple backplate mesh positioned behind the Gaussian scene. By compositing using the original background image rather than the in-camera LED captures, we eliminate light distortions introduced when capturing the real in-camera image-based emitters. Consequently, high-resolution capture of the LED background during the filming process is no longer required, reducing hardware requirements and overall production costs. This provides flexibility, enabling directors to display any desired background imagery, creating the possibility of designing capture scenarios that enhance actor immersion \cite{matthews2023habit, bennett2025facilitating, chen2024impact}. 

As no existing multi-view datasets support this capture configuration, we introduce a new streamlined data collection protocol. 
We captured three data sets using this methodology, using both standard and non-standard VP setups, in collaboration with Partner-XX\footnote{A research hub for film production and VFX.}. Furthermore, due to the lack of relevant benchmarks, we also designed several proxy baselines inspired by both forward and deferred rendering techniques. 

To summarize, this paper contributes:
\begin{itemize}
    \item A geometry-free relighting pipeline for VP that requires no depth priors, auxiliary geometry, light transport simulation, or custom CUDA/GPU code.
    
    \item A novel 3D GS parameterization that directly samples a known image-based light source to model photorealistic, view-dependent lighting without environment maps.

    \item The first publicly available multi-camera VP dataset for 3D 
    reconstruction and relighting, alongside an efficient capture protocol.
\end{itemize}

Our approach outperforms all baselines in both objective and perceptual quality, while remaining lightweight ($<$3\,GB RAM, $<$5\,GB VRAM), fast to train ($<$2 hours on $18\times$ 1080p images), and capable of rendering useful AOVs including depth, XYZ, lighting residual, and unlit renders.

\noindent All supplementary information is provided in the appendix and online: \url{RELEASED-UPON-ACCEPTANCE}.

\section{Related Work}\label{sec: related work}
\noindent \textbf{Environment Matting} historically aims to separate foreground imagery from background imagery by predicting per-pixel alpha values to produce a foreground RGBA image. Certain approaches also focus on 2D image relighting based on new lighting conditions \cite{peers2009compressive, wang2009kernel, sengupta2021light, chuang2000environment, matusik2002image, blinn1976texture}. Some specifically investigate matting and relighting for a capture setting where the clean plate is an image-based lighting source with known imagery \cite{chuang2000environment, peers2009compressive}. For example, after matting, \cite{chuang2000environment} proposes to model an implicit relationship between foreground pixels and pixels in the background image texture. This operates without BRDF-based shading by modeling a lighting intensity and background UV coordinate value per foreground pixel.


These methods often rely on specialized light-scanning equipment \cite{chuang2000environment, matusik2002image} and, because they operate in 2D, only support compositing pre-captured footage. In this paper, we extend the matting and relighting problem to include 3D reconstruction. This broadens creative possibilities, by allowing for novel view synthesis alongside matting and relighting. 

\noindent \textbf{Inverse Rendering} aims to decompose surface geometry and material properties using neural 3D reconstruction, to enable editing materials, textures and/or lighting post-capture \cite{gao2024relightable,liang2024gs,guo2024prtgs,zhu2025gaussian,ye2025geosplatting,sun2025generalizable,jiang2024gaussianshader,moenne20243d, bi2024gs3,wu2025deferredgs,chen2024gi, jin2023tensoir, kaleta2025lumigauss, yao2024reflective}. While these tackle a broad range of problems for different capture settings, they share the same underlying objective of using physically-based rendering to simulate light interacting with neural volumes. Methods operating under image-based illumination settings \cite{gao2024relightable, liang2024gs, guo2024prtgs, zhu2025gaussian, chen2024gi, wu2025deferredgs, ye2025geosplatting} generally focus on reconstructing a volumetric scene with a comprehensive set of surface and material properties while also reconstructing a 2D environment map. During rendering, the surface and material properties are used to determine view-dependent ray-based reflections, based on the viewport's pose and the estimated surface position and normals. The reflected rays are used to sample the environment map therefore, robust surface positions and normals are critical for accurate 3D reconstruction and relighting.

GS-IR~\cite{liang2024gs} achieves this by introducing a material parameterization for each Gaussian primitive that enables BRDF-based shading with a learnable environment map - jointly optimized with 3D reconstruction. To obtain geometric cues for shading, GS-IR applies regularization to stabilize the predicted normals during training. DeferredGS~\cite{wu2025deferredgs} follows a similar objective, where neural Gaussian parameters such as albedo, specular albedo, roughness, position and normals are rendered into intermediate buffers that are combined to compute shading from an environment map. To optimize the predicted normals, DeferredGS learns a 3D signed distance field and distills its gradients into the Gaussian representation. However, as emphasized by \citet{bi2024gs3}, prior works rely on geometric priors and/or aggressive regularization to learn accurate surface normals. This makes it challenging to accomplish relighting without either dense multi-view observations or reliable depth and normal priors. Instead, \citet{bi2024gs3} ($\text{GS}^3$ ) proposes a GS parameterization that embeds lighting intensity and view-dependent specular effects as per-Gaussian properties. The base colors are rendered into a shading buffer that is summed with a shadow and lighting map to produce the final render. However, this only applies to scenes lit by point lights so is not applicable to image-based illumination settings.

Although environment maps are commonly used, their formulation assumes far-field lighting. This introduces practical challenges in capture settings where illumination originates from nearby emitters such as LED walls. In practice, 360° cameras struggle to properly expose LED panels and introduce spatial distortions depending on camera placement. This makes it difficult to edit the predicted environment map, as it would require learning an un-distorted and un-exposed representation of the original background texture as well. This remains challenging for existing IR approaches. Furthermore, environment map parameterizations compress scene illumination into a spherical representation, which can reduce the resolution of reflections originating from structured light sources such as LED walls. Because many IR approaches already require substantial compute, modeling high-resolution environment maps also introduces additional optimization and memory challenges. Various methods have been proposed to handle the computational issues linked to ray-based shading, though these approaches require custom GPU code that limits portability.

A further limitation of many IR approaches \cite{gao2024relightable,liang2024gs,guo2024prtgs,zhu2025gaussian,ye2025geosplatting,sun2025generalizable,jiang2024gaussianshader,wu2025deferredgs,chen2024gi, jin2023tensoir, yao2024reflective} is that they assume single illumination conditions during training. In contrast, our capture setting observes the same scene under multiple image-based lighting conditions produced by different LED wall backgrounds. This relates more to work on 3D reconstruction ``in-the-wild'' (discussed in the next part). Among IR methods, \citet{jin2023tensoir} (TensoIR) proposes a solution for multi-illumination capture. TensoIR works by jointly reconstructing scene geometry and learning a separate environment map for each observable lighting condition. This introduces significant computational overhead, especially for footage capturing many lighting changes.

Ultimately, the reliance on environment maps arises from the assumption that scene illumination is unknown and external to the captured scene. However, in VP the illumination source is known, controlled through the LED wall imagery, and appears within the captured footage. Motivated by this, we directly sample the original background texture rather than reconstructing and sampling an environment map. This formulation is independent of the light source pose and avoids the need for ray-based shading, regularization and priors. Because our approach does not attempt to explicitly recover material properties, it significantly reduces the number of learnable parameters required for reconstruction and relighting. Our method can therefore be trained efficiently and deployed without custom GPU rendering frameworks, improving portability.

\noindent \textbf{3D reconstruction in-the-wild} aims to recover scene geometry and appearance from unconstrained photo collections, like landmark datasets exhibiting varied illumination and geometry \cite{martin2021nerf, li2025sparsegs, fridovich2023k, zhang2024gaussian, kulhanek2024wildgaussians, kaleta2025lumigauss}. These methods seek to model transient scene appearance as a function of global environmental changes, such as time of day or seasonal variation.

\citet{martin2021nerf} (NeRF-W) addresses this challenge by learning per-image appearance and transient embeddings that are decoded through a shallow MLP to predict static and transient volumetric properties. NeRF-W also learns a volumetric uncertainty term to suppress image-specific occluders and transient artifacts (e.g., pedestrians or vehicles) which appear in photo tourism datasets. \citet{fridovich2023k} (K-Planes) proposes a more explicit representation that learns per-image feature embeddings. These embeddings are decoded via a shallow MLP to approximate a learned color basis, which is combined with an appearance-independent volumetric feature to estimate the final color.

In-the-wild approaches therefore focus on modeling transient geometry and appearance through global illumination changes. Some methods even attempt to solve this problem with IR \cite{kaleta2025lumigauss, zhang2025rgs, feng2026uv, corona2026r3gw}. In contrast, our capture setup exhibits spatially localized appearance responses caused by controlled and local changes in the LED wall's imagery. This assumes that surfaces primarily respond to nearby regions of the LED wall content, while illumination changes in distant regions have negligible effect. In-the-wild methods are also designed for unconstrained capture scenarios, where illumination conditions are unknown and transient occluders are common. In our VP setting, however, lighting is controlled and scene geometry remains static throughout capture. This distinction motivates our objective of developing a geometry-independent relighting approach tailored to VP pipelines. 

\section{Gaussian Splatting}\label{sec: gs background}
The original 3D-GS model \cite{kerbl20233d} offers an explicit point-based approach to 3D reconstruction. Each primitive is assigned five learnable parameters, position $x_i \in \mathbb{R}^{M\times 3}$, scale $s_i \in \mathbb{R}^{M\times 3}$, quaternion rotation $r_i \in \mathbb{R}^{M\times 4}$, opacity $\sigma_i \in \mathbb{R}^M$ and color $c_i \in \mathbb{R}^{M\times 16\times 3}$. A 3D Gaussian with covariance $\Sigma_i = R_iS_iS_i^TR_i^T$  is projected onto the rendering camera's image plane, where $R_i\in \mathbb{R}^{3 \times 3}$ is the matrix representation of the quaternion $r_i$ and $S_i \in \mathbb{R}^{3 \times 3}$ is a diagonal matrix for $s_i$. The ``splatted'' Gaussians are then depth-sorted w.r.t. each pixel patch. To determine the weight of each Gaussian on every pixel $w_i(P)$, the Mahalanobis distance function,
\begin{equation}\label{eq: mahalanobis}
w_i(P) = \exp\!\left(
-\tfrac{1}{2}(P - X_i)^{\top} \Sigma_i^{-1} (P - X_i)
\right)
\end{equation}
is used based on the pixel's 2D projected center $P$ and the Gaussian's 2D projected center $X_i$. Splats with minimal impact are culled, and the remaining splats are alpha-blended in depth-order, forming the rasterized image $I_{j}$ for camera $j\in J$.

This paper focuses on the color parameter, $c_i$. This is represented using spherical harmonic coefficients, where an RGB signal is sampled using the camera's viewing direction $d_j$, defined as $\tilde{c_i} \in \mathbb{R}^{M \times 3}$. In this paper, the tilde identifies spherical harmonics components that have been transformed into RGB colors based on view direction.

\begin{figure*}
    \centering
    \includegraphics[width=\linewidth]{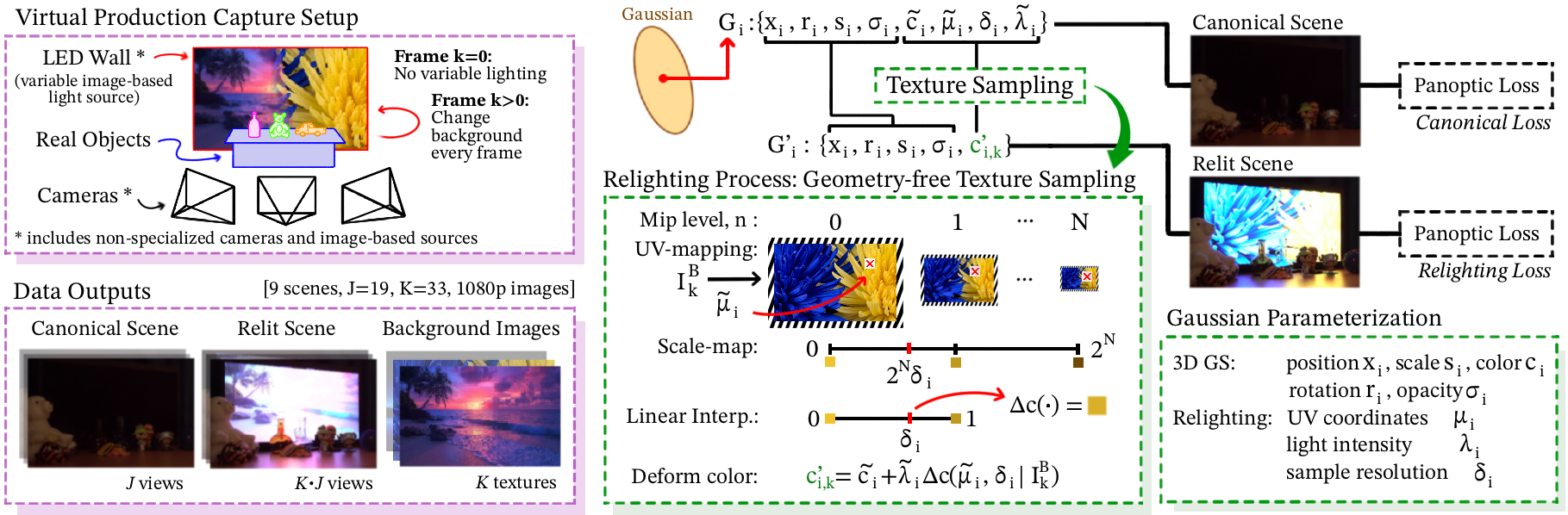}
    \caption{\textit{Left/Purple}: The proposed data collection setup and resulting datasets using professional miniature and life-size VP stages. \textit{Right/Green}: The proposed Gaussian  parameterization and relighting process. We learn three additional relighting parameters - used to sample a mipmap representation of the known background image. The canonical scene (fixed appearance) and relit scene (variable lighting) are then jointly trained.
    }
    \label{fig: main diagram}
\end{figure*}
\section{Method}\label{sec: main method}
Our proposed pipeline is shown in Fig.~\ref{fig: main diagram}, where we re-parameterize the Gaussian primitive with three additional parameters responsible for learning the lighting changes. In Sec.~\ref{sec: color deformation}, we present the rendering process and Sec.~\ref{sec: texture sampling} presents the strategy for sampling the image-based light source. Sec.~\ref{sec: ibl representation} presents the post-training method for rendering the background imagery. Finally, Sec.~\ref{sec: optimization} presents the optimization strategy.
 
\subsection{Rendering Process}\label{sec: color deformation}
Our approach to rendering relies on a parametric deformation function \cite{fridovich2023k, wu20244d, chao2025textured}. Our goal is to deform the color parameter based on the input background texture with 
\begin{equation}\label{eq: residual color}
    c_{i,k} = \tilde{c_i} + f(\cdot | I^B_{k}),
\end{equation}
\noindent where $f(\cdot|I^B_k)$ estimates the residual color caused by variable lighting, based on some inputs given the background texture $I^B_k$, and the final color is rendered using the vanilla 3D-GS rasterization \cite{kerbl20233d}. This resembles the BRDF shading function, where $\tilde{c_i}$ could be considered the diffuse color and the residual as outgoing light per source. However, this requires a dense set of parameters, as with IR (Sec.~\ref{sec: related work}).

Instead, we encourage $\tilde{c_i}$ to represent the canonical footage through the canonical loss, shown in Fig.~\ref{fig: main diagram} and discussed in Sec.~\ref{sec: optimization}. Hence, $f(\cdot | I^B_k)$ is responsible for variable lighting and shadowing. This re-interprets $\tilde{c_i}$ as the object color with fixed lighting and $f(\cdot | I^B_k)$ as the variable lighting color, where lighting changes are isolated to changes in $I^B_k$. In practice, this allows users to select the image-based light sources they wish to change while preserving the appearance from lighting they do not wish to change.

The function $f(\cdot|I^B_k)$ is decomposed into two components, lighting intensity $\tilde\lambda_i$ and change in color $\Delta c(\cdot | I^B_k)$. Hence,
\begin{equation}\label{eq: residual function}
    f(\cdot|I^B_k) = \tilde\lambda_i \Delta c(\cdot | I^B_k),
\end{equation}
\noindent where $\lambda_i \in\mathbb{R}^{M\times 16 \times 1}$ such that $\tilde\lambda_i \in \mathbb{R}^{M \times 1}$, and $\Delta c(\cdot | I^B_k) \in \mathbb{R}^{M \times 3}$ is presented in the following subsection. Note, we label $\tilde\lambda_i$ as light intensity for simplicity, though it is responsible for estimating both the LED's and camera's exposure settings as well as the intensity of outgoing light. In scenarios where outgoing rays are not expected to intersect the lighting source, we expect $\tilde\lambda_i \rightarrow 0$ such that $c'_{i,k} = c_i$, effectively producing an unlit surface. This is shown later in Fig.~\ref{fig: light intensity visual}, where $\tilde\lambda_i$ handles view-dependent refractions for glass objects.

\subsection{Geometry-Independent Light Texture Sampling}\label{sec: texture sampling}
The function $\Delta c(\cdot \mid I^B_k)$ models the view-dependent lighting variation for each Gaussian conditioned on the background image $I^B_k$. To enable geometry-free relighting, each Gaussian is parameterized by learnable UV coordinates\footnote{i.e., 2D pixel coordinates.} $\mu_i$, where $\mu_i \in \mathbb{R}^{M \times 16 \times 2}$ and $\tilde{\mu}_i \in \mathbb{R}^{M \times 2}$. These coordinates are used to bilinearly sample $I^B_k$ and directly correspond to image-space locations in the background texture where reflected or refracted rays are expected to intersect.

Naive sampling of high-resolution images introduces texture aliasing, a common problem in graphics that can be resolved using mipmaps \cite{williams1983pyramidal, heckbert1986fundamentals}.

Accordingly, we first downsample the original background image into $N$ levels, $I^B_{k,n} \in \mathbb{R}^{2^{-n}H, 2^{-n}W, 3}$ where $n \in [0,N)$. We then introduce a scaling parameter, $0 \leq \delta_i \leq 1$ where $\delta_i \in \mathbb{R}^{M\times 1}$, and derive an upper and a lower bound mip-level using $n_i^{LB} = \lfloor(2^N\delta_i)\rfloor$ and $n_i^{UB}= \lceil(2^N\delta_i)\rceil$. Per-Gaussian, a lower and an upper bound mipmap $\{I^B_{n_i^{LB}},I^B_{n_i^{UB}}\}$ are sampled to produce lower and upper bound color samples $c_{\text{LB}}(\tilde\mu_i, I^B_{n_i^{LB}})$ and $c_{\text{UB}}(\tilde\mu_i, I^B_{n_i^{UB}})$. The final color residual is acquired by linearly interpolating these samples using Eq.~\ref{eq: mipmap}, as illustrated in Fig.~\ref{fig: main diagram}.
\begin{equation}\label{eq: mipmap}
\Delta c (\tilde\mu_i, \delta_i |  I_k^B)
    = \alpha\, c_{\text{UB}}(\cdot)
    + (1-\alpha)\, c_{\text{LB}}(\cdot),
\end{equation}
\[
\text{where } 
\alpha = \frac{2^N \delta_i - 2^{n_i^{lB}}}{2^{n_i^{UB}} - 2^{n_i^{lB}}}.
\]

By modeling $\tilde\mu_i$ with spherical harmonics, our method learns view-dependent samples of the LED wall, implicitly encoding the intersection of a viewport ray reflect off an object surface. Doing this w.r.t. the LED wall's 2D image space avoids the need to compute surface normals and the LED wall's pose, allowing us to capture reflections and refractions with a single parameter.

\subsection{Representing and Rendering Virtual Backgrounds}\label{sec: ibl representation}
The perceptual quality of the reconstructed image-based light source is important for meeting the needs of a film production; however, GS is not capable of reliably reconstructing image-based light sources. To address this, we represent the virtual LED wall as a low-polygon 2D mesh. As light-texture sampling is geometry-independent, precise pose alignment is not required during optimization and can instead be performed manually as a post-process. Our code offers interactive tools for efficient manual posing.

We render the background mesh using a lightweight ray-intersection pass and alpha-blend it with the foreground GS render. Specifically, we cull Gaussians that appear within masks identifying the background imagery, with $Q_j$. These were manually drawn, for fine-grained control, but options exist to automate the segmentation process \cite{carion2025sam, agarwala2004keyframe, li2016roto,adobe_after_effects,nuke_foundry,blender_foundation}.

To reduce noise during training, we also regularize the Gaussian's alpha $I_{\alpha}^j$ using a mask loss, $L_{\text{mask}}=|I_{\alpha}^{j} - Q_j|$. At inference, the mesh and GS images are composited via alpha-blended using $I_{\alpha}^j$ to form the final render.

\subsection{Optimization}\label{sec: optimization}
\textit{Initialization.} Similar to prior works on IR \cite{gao2024relightable,liang2024gs,guo2024prtgs,zhu2025gaussian,ye2025geosplatting,sun2025generalizable,jiang2024gaussianshader}, we initialize our GS representation by pre-training a generic GS model for $x_i$, $r_i$, $s_i$, $c_i$, and $\sigma_i$, using the canonical ground truth footage $I^*_j$. We choose to initialize the zero-th degree coefficients of $\lambda_i$ to 0.01 and the remaining coefficients to zero. Similarly, the zero-th $\mu_i$ coefficients are set to 0.5 (the center of the texture) and the remainder is set to zero. Then, $\delta_i$ is initialized with small random noise where $\delta_i = 0.99 + \epsilon$. Finally, initial points that sample $Q_j (X_i) = 0$ are culled, (i.e.) when the 2D means $X_i$ intersect the ground truth mask.

\textit{Adaptive Density Control.} For gaussian densification, we employ the vanilla 3D-GS scheme in \cite{kerbl20233d}.

\textit{Training Scheme.} The training scheme relies solely on the panoptic loss,
\begin{equation}
\text{Loss} = (1-\gamma_{\text{DSSIM}})L_{\text{RGB}} + \gamma_{\text{DSSIM}}L_{\text{DSSIM}} + \gamma_{\text{canon}}L_{\text{canon}},
\end{equation}
where $L_{\text{RGB}} = |I'_{j, k} - Q_j \cdot I^*_{j, k}|$,
and $L_{\text{canon}} = |I_{j} - Q_j \cdot I^*_{j} |$.
Here, $I'_{j,k}$ is the image rendered from camera $j$ under the $k$-th background texture; $I^*$ denotes ground truth. $L_{\text{DSSIM}}$ is the DSSIM analogue of $L_{\text{RGB}}$, and $\gamma_{\text{DSSIM}} = \gamma_{\text{canon}} = 0.2$. 

\section{Data Capture and Datasets}\label{sec: data}
\paragraph{Capture Setup.} The data collection setup involves capturing $J$ fixed views of a scene under $K$ varying illuminations as well as $J$ fixed views without the variable illumination; shown in Fig.~\ref{fig: main diagram}. We accomplish this with 19 Sony A7 II cameras capturing 1080p resolution; 18 views are used for training and 1 view is used for testing, following precedent set by datasets on sparse-view 2.5-D scene reconstruction \cite{li2022neural}. The sparse-view capture constraints simulate the case when camera budgets are tight and dense multi-view capture is not possible.  

\paragraph{Datasets.} Three datasets were captured with 99 different background lighting conditions. Each dataset is split into three scenes, where $K=23$ backgrounds are used for training and $K=10$ for testing. Dataset 1 uses an LCD monitor as the primary light source and simulates a miniature VP stage. In collaboration with Partner-XX, Dataset 2 and 3 were captured using a professional VP set with large LED walls acting as backlighting. Datasets 1 and 2 capture a broad range of materials including but not limited to metal, plastic, glass, wood and fluff. Dataset 3 captures a diverse set of human clothing. All objects vary in shape, texture and size as shown in Fig.~\ref{fig: main results}. Finally, the background masks, $Q_j$, were manually drawn and are used during testing to avoid evaluating light sources.

\paragraph{Camera calibration and GS initialization.}
We use Nerfstudio \cite{nerfstudio} to generate poses and intrinsics, and use Splatfacto with default hyper-parameter settings to train an initial scene; this takes $<10$ minutes. For sparse multi-view datasets, additional views may be required for accurate posing. We suggest capturing additional views of the canonical scene and using this data for camera calibration and initialization. This was done for Dataset 1 but not for Dataset 2 or 3.

\section{Experimental Results}\label{sec: experiments}
This section is split into three parts: (i) objective results, (ii) a visual assessment of localized lighting patterns, and (iii) a subjective study that assesses the potential impact of our work in practice.

The quantitative experiments test novel view and lighting synthesis capabilities. We use RGB-based PSNR, SSIM and LPIPS-VGG, and separately compute PSNR for Y and CrCb channels; $\text{PSNR}_Y$ is especially useful for assessing texture accuracy. All experiments use an Nvidia H100 GPU for training and an RTX 3090 for inference. All methods were tuned with a single hyperparameter search across all datasets. The full set of metric and video results are available in the appendix and online.

\subsection{Quantitative Experiments}\label{sec: quantitave subsec}
\begin{table*}[t]
    \centering
    \caption{Results on the novel view and lighting conditions. Averages are provided per dataset, using three-fold cross validation.}
    \resizebox{0.8\linewidth}{!}{%
    \begin{tabular}{c c c c c c c}
    \toprule
    Data & Method & $\text{PSNR}_{\text{RGB}}$ & $\text{PSNR}_{\text{Y}}$ & $\text{PSNR}_{\text{CrCb}}$ & SSIM & LPIPS \\ 
    \midrule
    
    Set 1
    & TensoIR & 16.19 & 16.81 & 28.04 & 0.655 & 0.591  \\
    & TIR & 23.20 & 23.95 & 33.84 & 0.895 & 0.300 \\
    & DFR & 22.04 & 22.74 & 32.44 & 0.863 & 0.318 \\
    & Ours & \textbf{25.65} & \textbf{26.38} & \textbf{35.87} & \textbf{0.904} & \textbf{0.290} \\
    
    \midrule
    
    Set 2 
    & TensoIR & 15.67 & 16.84 & 25.78 & 0.766 & 0.255  \\
    & TIR & 19.44 & 20.77 & 28.04 & 0.809 & 0.194 \\
    & DFR & 18.84 & 20.13 & 27.51 & 0.784 & 0.224 \\
    & Ours & \textbf{22.03} & \textbf{23.82} & \textbf{29.75} & \textbf{0.839} & \textbf{0.187} \\
    
    \midrule
    
    Set 3 
    & TensoIR & 18.19 & 19.76 & 27.45 & 0.618 & 0.475 \\
    & TIR & 23.04 & 24.49 & 32.29 & 0.930 & 0.175 \\
    & DFR & 20.69 & 22.00 & 30.87 & 0.895 & 0.219 \\
    & Ours & \textbf{24.97} & \textbf{26.13} & \textbf{33.85} & \textbf{0.936} & \textbf{0.166} \\
    
    \bottomrule
\end{tabular}
    }
    \begin{tabular}{c c c c c}
            \multicolumn{5}{c}{Average Statistics} \\ \toprule
            & Training Speed  & RAM & VRAM & Inference \\ 
            &(minutes)  & (GB) & (GB) & (FPS) \\ \hline
            TensoIR & 296 & 30.5 & 10.0 & 0.02\\
            TIR & 103 & 2.5 & 5.0 & 28 \\
            DFR &  42 & 2.5 & 4.1 & 38 \\
            Ours & 112 & 2.5 & 4.4 & 35\\ 
            \bottomrule
    \end{tabular}
    \label{tab: main results}
\end{table*}

\subsubsection{Performance comparison}\label{sec: baseline experiments}
As discussed in Sec.~\ref{sec: related work}, few works investigate both multi-view and multi-illumination problems. We are only aware of TensoIR, for which benchmarks on multi-illumination are absent \cite{jin2023tensoir}. Following precedent set \cite{furukawa2009accurate, barron2014shape, ren2013global}, we introduce two additional proxy baselines to closely assess the parametrization and texture sampling process. The first proxy baseline adapts TensoIR to a geometry-free pipeline and evaluates the case when a triplane representation \cite{jin2023tensoir, fridovich2023k} learns $\lambda_i$ and $\mu_i$, by decoding spatially variant shared features; named TIR after TensoIR. The second proxy baseline applys a deferred rendering approach to a geometry-free pipeline, and adapts the differentiable splatting process in \cite{liang2024gs} to produce image-buffers for $c_i, \lambda_i, \mu_i \rightarrow I_{c},I_{\lambda}, I_{\mu}$. These are used to deform the canonical signal with $I'_{j,k} = I^{j}_c + I_{\lambda}^{j} \cdot I_{\Delta c}^{j,k}(I^{j}_{\mu}, I^B_k)$. This is named DFR after deferred rendering. More information and diagrams are available in the appendix.



\begin{figure}
    \centering
    \includegraphics[width=\linewidth]{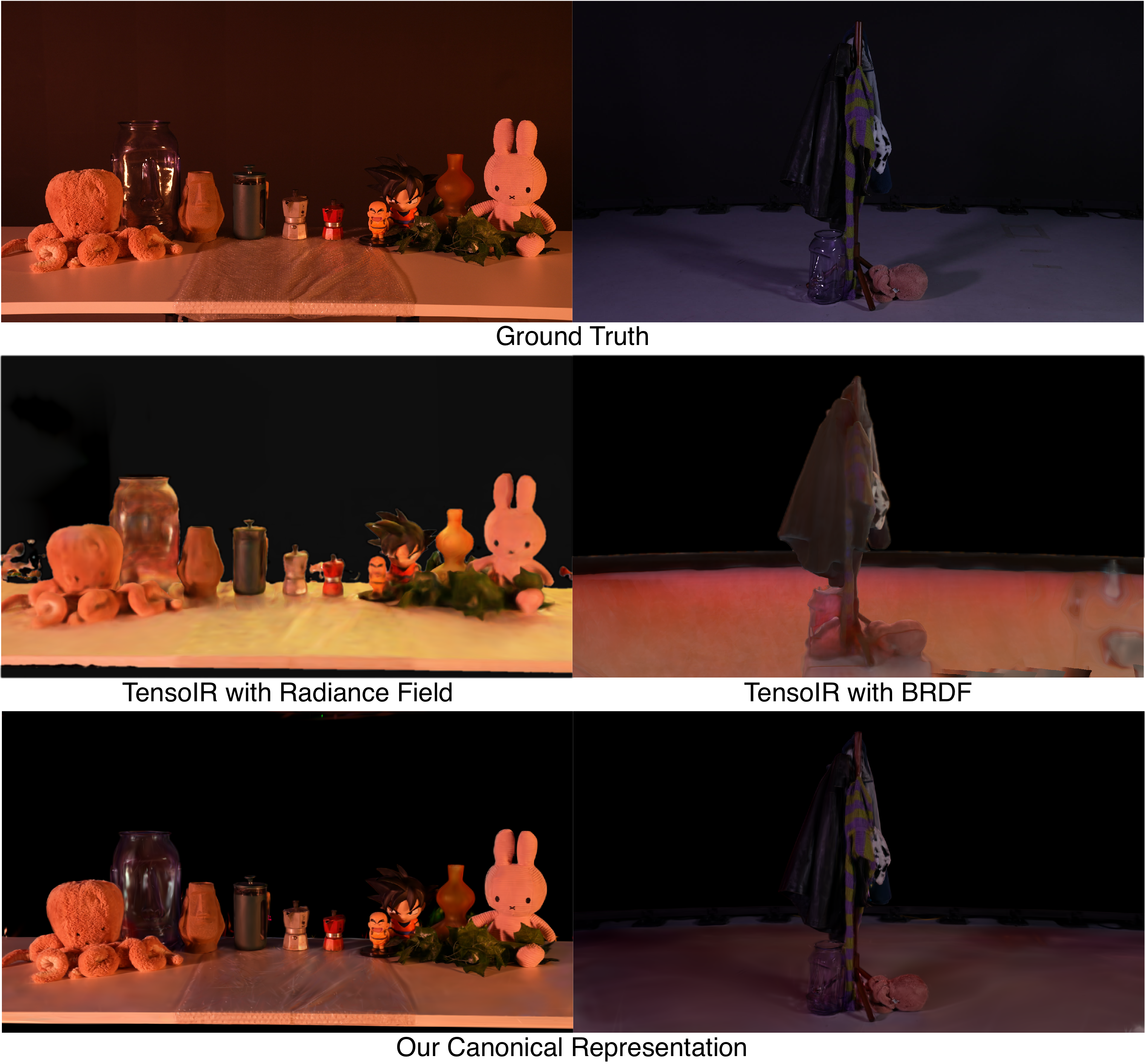}
    \caption{Comparing unlit/canonical scenes between the Ground Truth, TensoIR and Our method. \textit{TensoIR with Radiance Field} is the canonical reference without BRDF. \textit{TensoIR with BRDF} is the canonical reference when the BRDF is used but entirely black (no environmental light).}

    \label{fig: canon comparison} 
\end{figure}

Tab.~\ref{tab: main results} and Fig.~\ref{fig: main results} confirm that our framework produces the best metric and visual results. In contrast, TensoIR performs poorly on relighting tasks, as the scene geometry and learned environment maps lack sufficient resolution and detail. This is also shown in Fig.~\ref{fig: canon comparison} when comparing the canonical/unlit results between TensoIR and Our pipeline. In comparison, our framework eliminates the need to learn environment maps, enabling more intuitive and direct control over lighting by directly modifying $I^B_k$.

DFR provides a significant boost in speed and compactness, but the overall quality is poor. TIR shows decent quality but requires more computation and is slower at inference. Our method, by contrast, strikes a balance with rendering speed and computation, and delivers high-quality results with a compact model.

\subsubsection{Ablation study}\label{sec: dataset ablation}
\begin{table*}[t]
    \centering
    \caption{Average reconstruction-only results: Comparing 3DGS (the default) and Mip-Splatting \cite{yu2024mip} as backbones for our framework.}
    \begin{tabular}{c c c c c}
        \toprule
            Method & $\text{PSNR}_{\text{RGB}}$ & $\text{PSNR}_{\text{Y}}$ & $\text{PSNR}_{\text{CrCb}}$  & SSIM \\ \hline
            3DGS & 25.62 & 26.11 & 37.27 & 0.893 \\
            Ours + 3DGS & 25.77 & 26.49 & 36.29 & 0.892 \\ \hline
            Mip-Splatting & 24.96 & 25.06 & 36.89 & 0.885 \\ 
            Ours + Mip-Splat. & 24.86 & 25.36 & 36.33 & 0.883 \\ \hline
    \end{tabular}
    \label{tab: reconstruction only results}
\end{table*}

\paragraph{Reconstruction-only results.}
Tab.~\ref{tab: reconstruction only results} compares our framework with 3DGS \cite{kerbl20233d} and Mip-Splatting \cite{yu2024mip} on reconstruction-only tasks. This shows that geometric quality is not impacted by our relighting process, and that our framework is adaptable to different GS representations. This highlights the benefit of our non-invasive, geometry-free relighting framework.

\begin{table*}[t]
    \centering
    \caption{Ablations on Dataset 3: (i) Varying $J$ cameras. (ii) Varying $K$ textures. (iii) Using (a) variable lighting reference for the canonical image $I^*_j = I^*_{j,k=1}$, (b) no canonical reference $\gamma_{\text{canon}}=0$}
    \begin{tabular}{c c c c c}
        \toprule
        Method & $\text{PSNR}_{\text{RGB}}$ &$\text{PSNR}_{\text{Y}}$ & $\text{PSNR}_{\text{CrCb}}$ &  SSIM  \\ \hline
        (i.a) $J=6$ & 20.83 & 21.34 & 33.05 & 0.926  \\
        (i.b) $J=12$& 24.56 & 25.66 & 33.47 & 0.931\\
        (i.c) $J=18$& 24.69 & 25.77 & 33.69 & 0.935 \\ \hline
        (ii.a) $K=1$& 20.25 & 21.86 & 27.89 & 0.906\\
        (ii.b) $K=10$& 23.24 & 24.27 & 32.72 & 0.928 \\
        (ii.c) $K=20$& 23.82 & 24.88 & 33.08 & 0.931 \\
        (ii.d) $K=30$& 24.53 & 25.48 & 34.05 & 0.938\\ \hline
        (iii.a) $I^*_j = I^*_{j,k}$ &  22.65 & 23.83 & 31.52 & 0.927\\ 
        (iii.b) $\gamma_{\text{canon}}=0$& 25.21 & 26.33 & 33.91 & 0.937\\
        \bottomrule
    \end{tabular}
    \label{tab: ablations}
\end{table*}

\begin{figure}
    \centering
    \includegraphics[width=\linewidth]{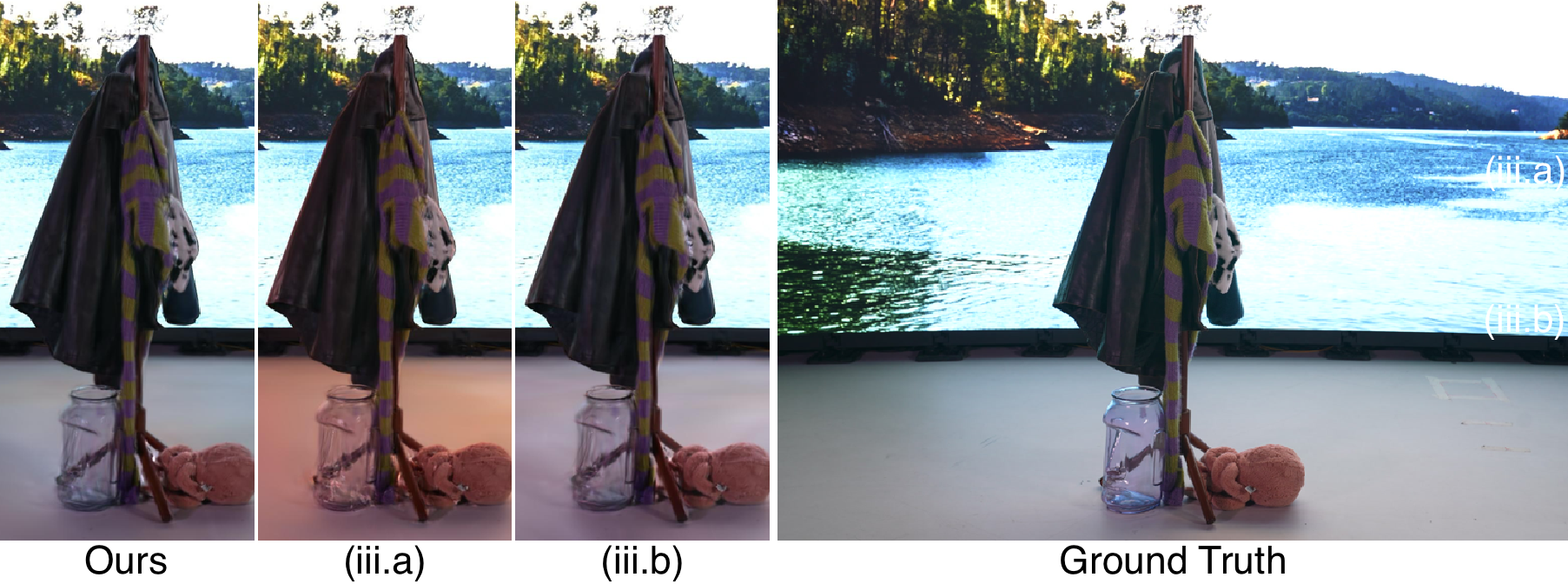}
    \caption{Visual comparison between our final renders and the ablations (iii).}
    \label{fig: ablationsiii}
\end{figure}

\paragraph{Dataset ablation.} The results are provided in Tab.~\ref{tab: ablations} and the ablations are described in the caption. Experiments (i-ii) indicate susceptibility to the number of textures and views used in training. The results of (i) are expected, though there may be benefit in employing sparse-view GS models \cite{fan2024instantsplat, zhang2024cor, chen2024mvsplat,liu2025review} to act as the backbone GS pipeline in future work. Experiments (ii) reveal a reliance on texture diversity. This leads us to hypothesize that a single background texture could be designed to optimize reconstruction and relighting quality when $K=1$; this is explored in the appendix. Finally, Tab.~\ref{tab: ablations} (iii) indicates that the proposed pipeline performs better without $L_{\text{canon}}$. Fig.~\ref{fig: ablationsiii} also demonstrates similar lighting results between Ours and (iii.b).   However, in practice (iii.b) prevents the canonical representation from preserving the fixed lighting setup. This is less useful for application as interpretability and controllability are important factors in any film-making workflow.

\subsection{Local Lighting Response to Local Lighting Changes.}
\begin{figure*}
    \centering
    \includegraphics[width=\linewidth]{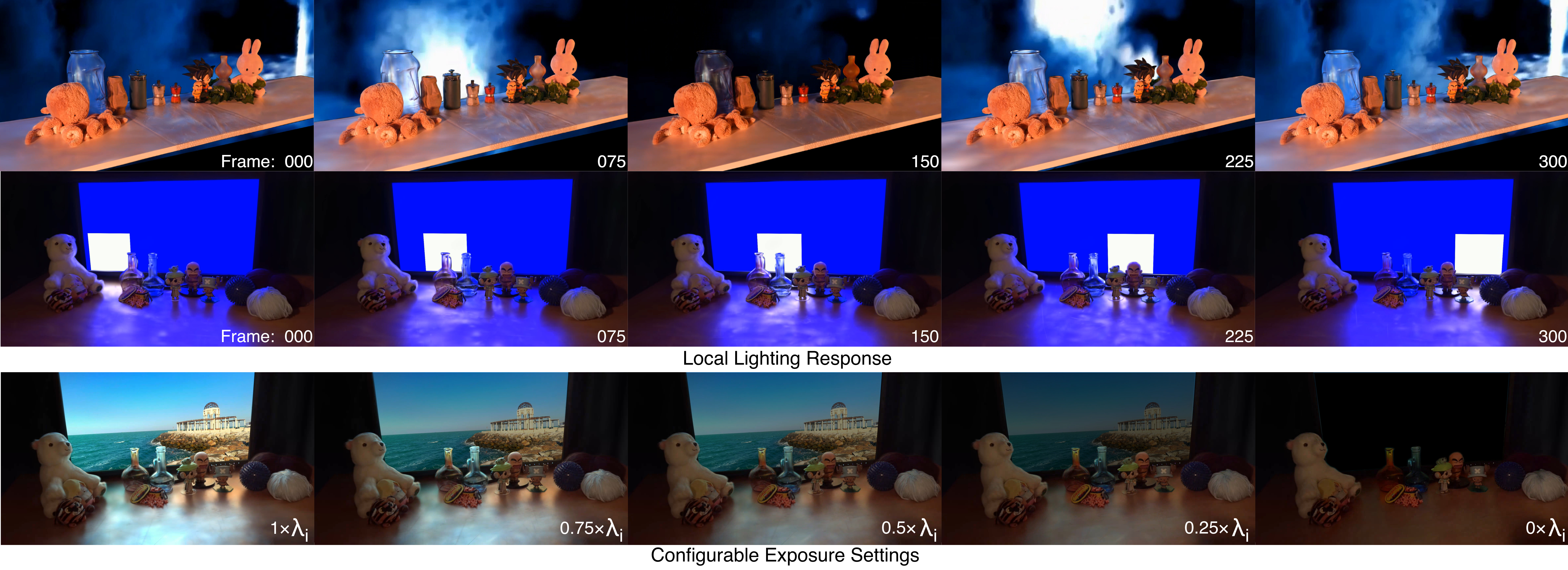}
    \caption{Visualizing localized response to dynamic image-based lighting textures as well as changes in lighting intensity.}
    \label{fig: dynamic textures} 
\end{figure*}

\begin{figure}
    \centering
    \includegraphics[width=\linewidth]{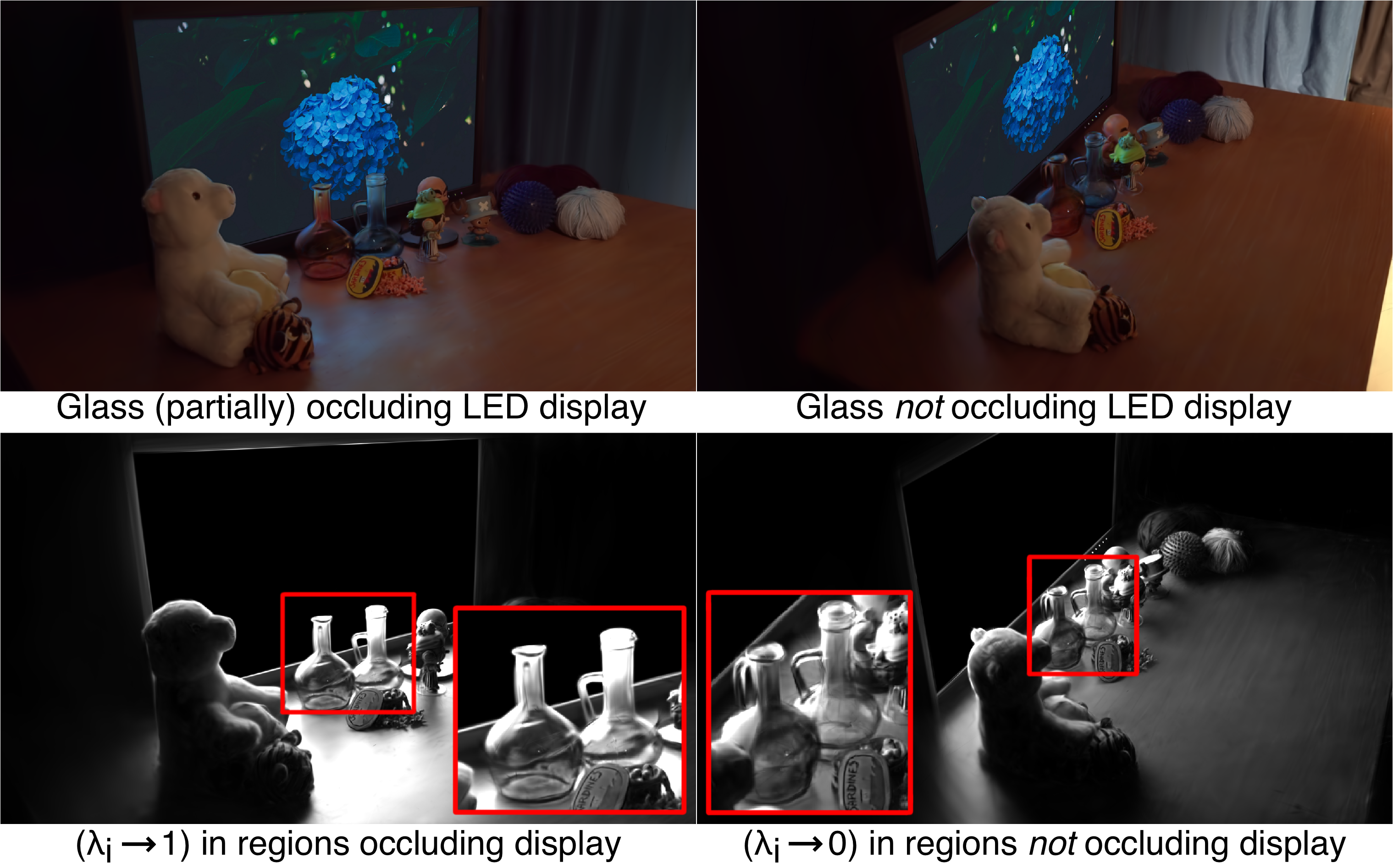}
    \caption{Demonstrating changes in $\lambda_i$ based on view-dependent changes where the top row renders the relit scene and the bottom row renders $\lambda_i$. Specifically, for refractive/transmissive objects: $\lambda_i \rightarrow 1$ when transparent objects occlude the image-based light source and $\lambda_i \rightarrow 0$ when it does not.}

    \label{fig: light intensity visual} 
\end{figure}

With dynamic backgrounds, we expect to find a physical relationship between the local changes in $I^B_k$ and the local GS lighting response. The same is true when modifying the exposure setting, determined by $\lambda_i$. This is demonstrated in Fig.~\ref{fig: dynamic textures} and~\ref{fig: light intensity visual}, where we observe the relit scene locally adapting to a variety of physical and visual changes. Notably, Fig.~\ref{fig: light intensity visual} shows $\tilde\lambda_i$ adapting accordingly for transparent/refractive texture.

\subsection{Subjective study}\label{sec: subjective study}
\begin{figure*}
    \centering
    \includegraphics[width=\linewidth]{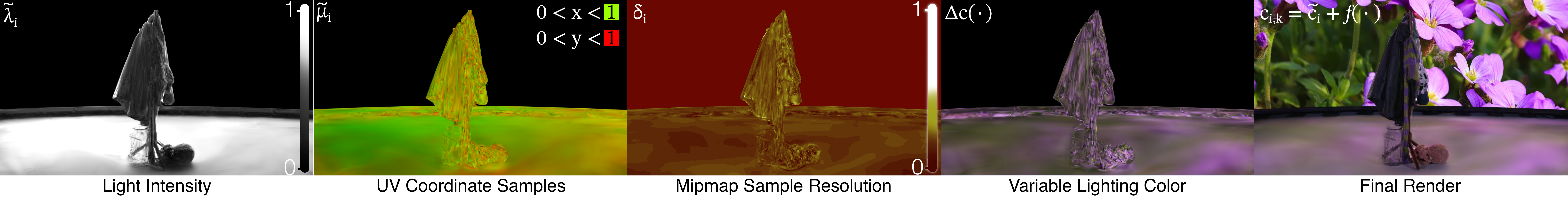}
    \caption{Additional AOVs synthesized by our pipeline for subjective assessment.}

    \label{fig: aovs} 
\end{figure*}

This section evaluates the impact of our pipeline in practical VFX workflows. Three artists from Partner-XX and Partner-YY\footnote{A research-focused VFX studio.} were tasked with performing single-image relighting on samples from Datasets 2 and 3, both with and without the AOVs synthesized by our pipeline. Their results were compared against the direct lighting predictions produced by our model. To simulate production constraints, each manual task was limited to one hour. Additional details are provided in the appendix and supplementary material.
\begin{enumerate}
    \item[A.] Manually relight a scene based on a target background.
    \item[B.] Same as (A) but with our additional relighting estimate and AOVs, shown in Fig.~\ref{fig: aovs}.
    \item[C.] Our pipeline's raw relighting predictions.
\end{enumerate}
We recruited 12 participants, including computer vision specialists and VFX artists, to evaluate the results. Five images were generated for each task (A–C). For each participant, three images per task were randomly selected and presented in random order. Participants evaluated each image based on three criteria: overall realism (whether objects appear realistic or artificial), physical lighting plausibility (whether shadows, highlights, and reflections are physically consistent), and foreground–background lighting consistency (whether the foreground matches the lighting conditions of the scene). Ratings were recorded using a 5-point Likert scale~\cite{likert1932technique}. 

\begin{figure}
    \centering
    \includegraphics[width=\linewidth]{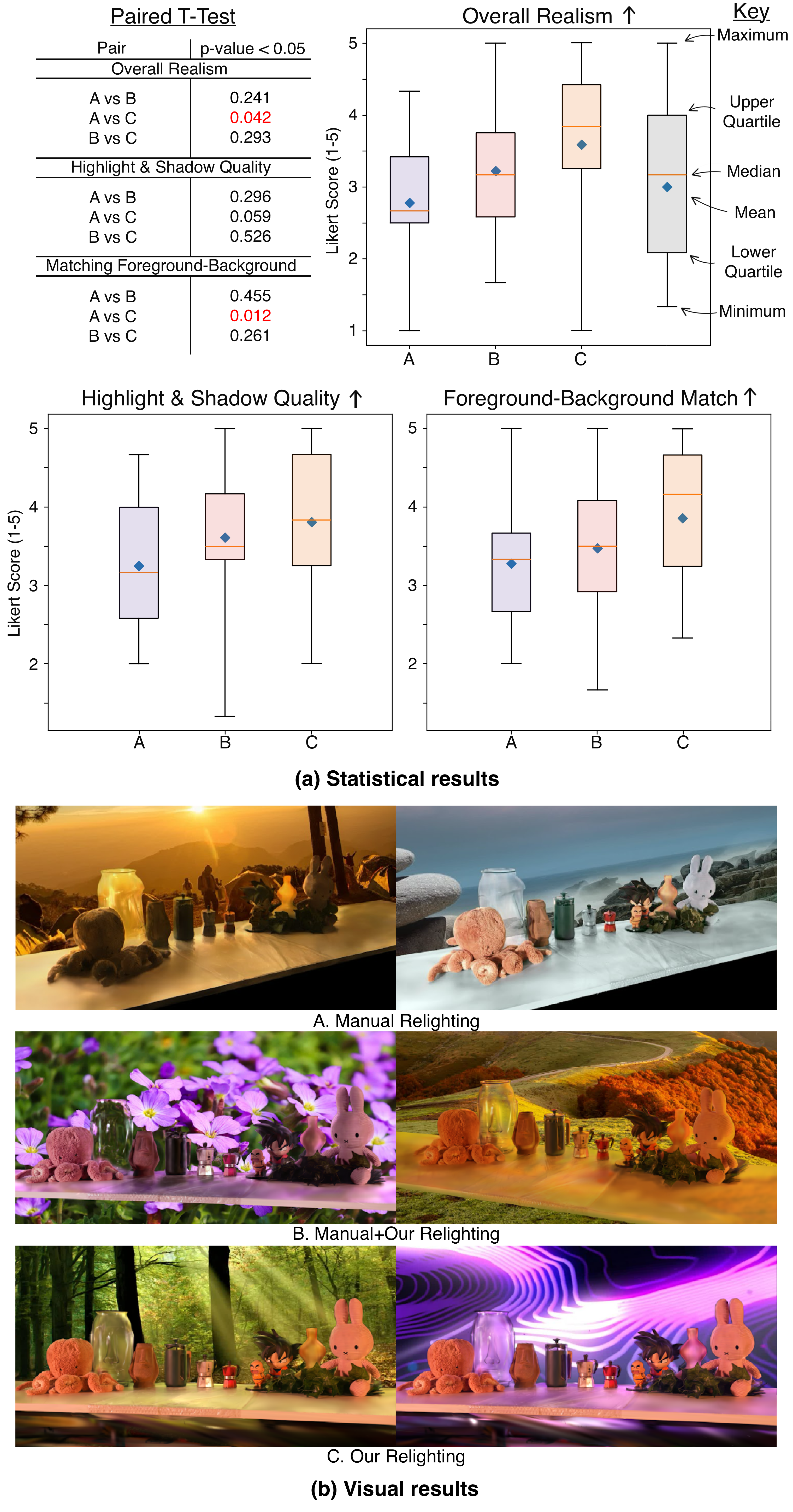}
    \caption{(a) \textit{Top-left}: Statistics on the paired t-test (red indicates statistically significant differences). \textit{Remainder}: Box plots of the three subjective conditions tested for tasks A-C and measured using the Likert scale where higher scores are better. (b) Examples of visual results used in the subjective study.}

    \label{fig: perceptual stats} 
\end{figure}

Fig.~\ref{fig: perceptual stats} shows the distribution of per-participant mean scores, along with paired t-test results for each task and criterion. Comparing tasks A and B, we observe a higher mean score for B, although the difference is not statistically significant. Partner-YY noted that task B enabled artists to focus on object-level edits, rather than low-level operations such as roto-scoping or histogram matching. However, the imposed time constraint limited the extent to which such edits could be applied across the full scene. As a result, task A exhibits a slightly tighter interquartile range (IQR), indicating more consistent outcomes, though again without statistical significance. A similar trend is observed when comparing tasks C and B, where task C shows a higher IQR. Overall, perceptual results and quantitative metrics (Sec.~\ref{sec: baseline experiments}) suggest that, under time-constrained conditions, our pipeline alone produces high-quality results without the additional overhead associated with manual workflows.
We expect this advantage to become more pronounced in more complex relighting scenarios, where manual approaches require substantially greater time and planning, while our method incurs no additional steps.

\section{Limitations and Future Work}\label{sec: future work}
\begin{figure*}
    \centering
    \includegraphics[width=\linewidth]{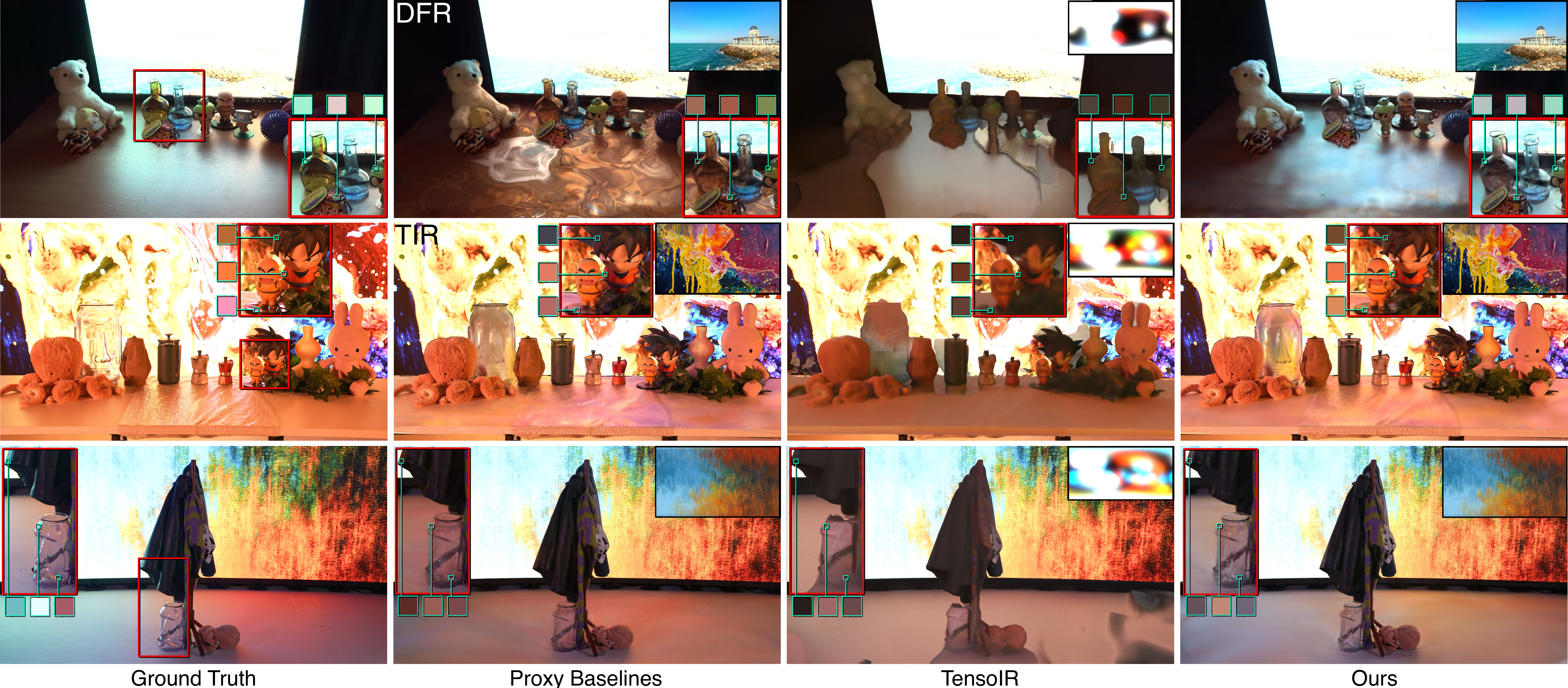}
    \caption{Comparing our method to baselines and ground truth images. Results for Dataset 1, 2 and 3 are shown in order of top-to-bottom. On the top-right inset of each render we show the image-based texture map sampled by each approach: TensoIR learns an low-resolution ($512 \times 256$ pixels) environment map while all other methods sample the (1080p) ground truth background texture $I^B_k$. We also show zoomed in results and color pick pixel patches with a high degree of light intensity to asses the quality of the colored reflections and refractions.}

    \label{fig: main results}
\end{figure*}

The primitive count in each scene is forcibly kept low, as more points would lead to a higher rate of over-reconstruction for sparse multi-view setups. For relighting, this primarily affects surfaces that exhibit low-frequency textures under canonical lighting but high-frequency textures after relighting, as shown by the glass head and table-top in Fig.~\ref{fig: main results}. We believe this is due to the reliance on the vanilla adaptive density control \cite{kerbl20233d}, so future work may benefit from schemes that boost densification rates for primitives in the aforementioned regions of the reference images. 

\section{Conclusion}\label{sec: conclusion}
This paper presents a method for reconstructing and relighting real-world 3D scenes under varying image-based lighting conditions. Targeting virtual production, our approach leverages a compact Gaussian Splatting representation that is both efficient to train and robust in practice. Crucially, our lighting formulation is geometry-free, avoiding the propagation of geometric inaccuracies inherent to Gaussian Splatting into the relighting process, as commonly observed in neural inverse rendering methods. We validate our approach against TensoIR and a set of proxy baselines, demonstrating consistent improvements in visual quality, compactness, and both training and inference efficiency. A user study further shows that our results are perceptually comparable to professional manual relighting, suggesting that artists can redirect their effort toward higher-level creative tasks.

To enable full compositing, we introduce a background replacement strategy in which the scene background is masked during training and reintroduced post-training as a low-polygon mesh. This mesh displays the original background image without lighting distortions present in the original footage, and is rendered using a lightweight rasterizer before being alpha-blended with the foreground Gaussian scene. By removing the dependence on capturing LED-based background imagery during training, our approach relaxes key virtual production constraints, including the need for specialized cameras and display setups. We demonstrate that high-quality reconstruction, relighting, and compositing can be achieved from footage captured with standard cameras and non-specialized lighting conditions, offering a pathway to significantly reduce production cost and complexity.

\newpage


\bibliographystyle{plainnat}
\bibliography{sample-base}

\end{document}